\documentclass[conference]{IEEEtran}
\usepackage{times}

\usepackage[sort&compress, numbers]{natbib}
\usepackage{multicol}
\usepackage[bookmarks=true]{hyperref}
\usepackage{graphics} 
\usepackage{graphicx}
\usepackage{epsfig}
\usepackage{amsmath,nccmath} 
\usepackage{amssymb}  
\usepackage{multirow}
\usepackage{booktabs}
\usepackage[table]{xcolor}
\usepackage{color,soul}
\usepackage{acronym}
\usepackage{mathrsfs}
\usepackage{nicefrac}
\usepackage{subfigure}

\usepackage{lipsum}

\usepackage{tikz}
\usetikzlibrary{calc, arrows, shapes, positioning}
\usepackage{pgfplots}

\definecolor{lightblue}{rgb}{36,109,226}

\newcommand{\change}[1]{{\color{black} #1}}

\newcommand{\bbm}{\begin{bmatrix}}
\newcommand{\ebm}{\end{bmatrix}}
\newcommand{\mbf}{\mathbf}

\newcommand{\mbs}[1]{{\boldsymbol{#1}}}

\newcommand{\beq}{\begin{equation}}
\newcommand{\eeq}{\end{equation}}
\newcommand{\bdis}{\begin{displaymath}}
\newcommand{\edis}{\end{displaymath}}
\newcommand{\beqn}[1]{\begin{subequations}\label{eq:#1}\begin{eqnarray}}
\newcommand{\eeqn}{\end{eqnarray}\end{subequations}}
\acrodef{ESGVI}{Exactly Sparse Gaussian Variational Inference}
\acrodef{ELBO}{Evidence Lower Bound}
\acrodef{EM}{Expectation-Maximization}
\acrodef{MAP}{Maximum A Posteriori}
\acrodef{WNOA}{white-noise-on-acceleration}
\acrodef{2D}{2-dimensional}
\acrodef{3D}{3-dimensional}
\acrodef{RBF}{Radial Basis Functions}
\acrodef{VAE}{Variational Autoencoder}
\acrodef{SLAM}{Simultaneous Localization and Mapping}
\acrodef{ICP}{Iterative Closest Point}
\acrodef{LOAM}{Lidar Odometry and Mapping}
\acrodef{SuMa}{Surfel-based Mapping}
\acrodef{HERO}{Hybrid-Estimate Radar Odometry}
\acrodef{GEM}{Generalized \ac{EM}}
\acrodef{SGD}{Stochastic Gradient Descent}

\makeatletter
\def\footnoterule{\kern-3\p@
	\hrule \@width 1.15in \kern 2.6\p@} 
\makeatother

\IEEEoverridecommandlockouts

\usepackage{etoolbox}
\makeatletter
\patchcmd{\@makecaption}
{\scshape}
{}
{}
{}
\makeatother

\begin{document}

\title{Radar Odometry Combining Probabilistic Estimation and Unsupervised Feature Learning\vspace{-3mm}}




%
\author{\authorblockN{Keenan Burnett*,
David J. Yoon*,
Angela P. Schoellig, and
Timothy D. Barfoot}
\authorblockA{Institute for Aerospace Studies, University of Toronto\\\vspace*{-9mm}} \thanks{* Equal contribution.}
}

\maketitle
\begin{abstract}This paper presents a radar odometry method that combines probabilistic trajectory estimation and deep learned features without needing groundtruth pose information. The feature network is trained unsupervised, using only the on-board radar data. With its theoretical foundation based on a data likelihood objective, our method leverages a deep network for processing rich radar data, and a non-differentiable classic estimator for probabilistic inference. We provide extensive experimental results on both the publicly available Oxford Radar RobotCar Dataset and an additional 100 km of driving collected in an urban setting. Our sliding-window implementation of radar odometry outperforms most hand-crafted methods and approaches the current state of the art without requiring a groundtruth trajectory for training. We also demonstrate the effectiveness of radar odometry under adverse weather conditions. Code for this project can be found at: {\scriptsize \url{https://github.com/utiasASRL/hero_radar_odometry}}
\end{abstract}

\vspace{-1.5mm}

\IEEEpeerreviewmaketitle

\section{Introduction}

\vspace{-1mm}While the reliability of autonomous vehicles continues to improve, operating in rain and snow remains a challenge. Most autonomous driving systems rely primarily on a combination of cameras and lidar for perception, with radar sensors taking a back-seat role \cite{burnett_jfr20}. \citet{cen_icra18} presented a successful application of a scanning radar sensor to large-scale outdoor ego-motion estimation. Their work has inspired a resurgence of research into radar-based perception and navigation systems. Compared to lidar, radar is more robust to precipitation due to its longer wavelength. For this reason, radar may be a key to enabling self-driving vehicles to operate in adverse weather. The ultimate goal of this research is to approach the performance of lidar-based algorithms in ideal conditions and surpass them in adverse conditions.

\vspace{-0.5mm}

Previous works in this area have made significant progress towards radar-based odometry \cite{cen_icra18, cen_icra19, aldera_icra19, aldera_itsc19, barnes_corl19, barnes_icra20, park_icra20, hong_iros20, burnett_ral21, adolfsson_arxiv21, kung_icra21} and place recognition \cite{kim_icra20, saftescu_icra20, gadd_plans20, demartini_sens20, tang_icra20}. However, previous approaches to radar odometry have either relied on hand-crafted feature extraction \cite{cen_icra18, cen_icra19, aldera_icra19, aldera_itsc19, burnett_ral21, hong_iros20, adolfsson_arxiv21, kung_icra21}, correlative scan matching \cite{park_icra20, barnes_corl19}, or a (self-)supervised learning algorithm \cite{barnes_corl19, barnes_icra20} that relies on trajectory groundtruth. Barnes and Posner \cite{barnes_icra20} previously showed that learned features have the potential to outperform hand-crafted features. On the other hand, it has not yet been shown whether correlative scan matching systems can be scaled up to large-scale mapping and localization. In order to address these limitations, we propose a method that is able to learn features directly from radar data without relying on groundtruth pose information.\vspace{-0.5mm}


\begin{figure}[ht]
  \includegraphics[width=0.95\columnwidth]{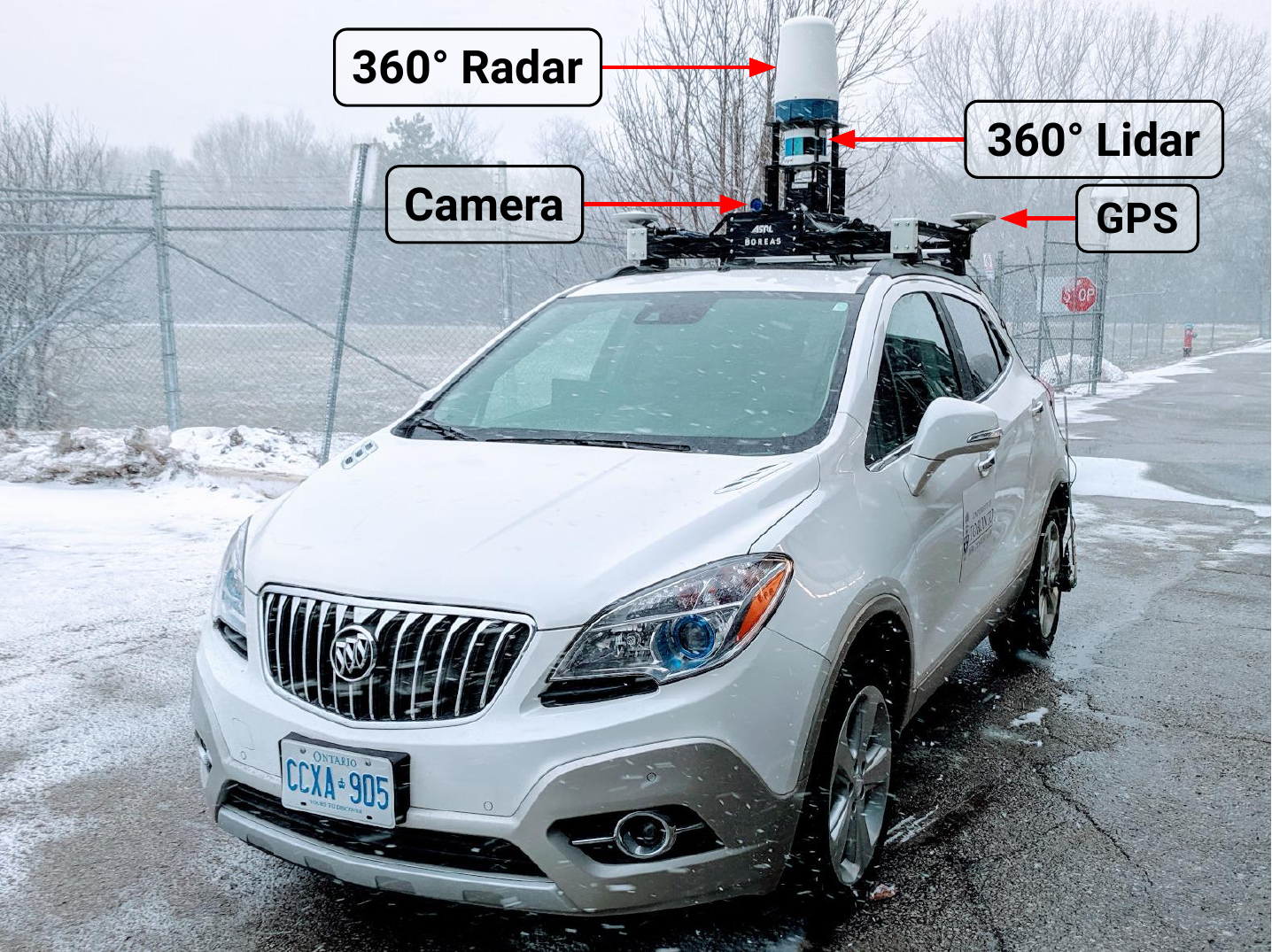}
  \centering
  \caption{Our data-taking platform, \textit{Boreas}, which includes a Velodyne Alpha-Prime (128-beam) lidar, Navtech CIR204-H radar, FLIR Blackfly~S monocular camera, and Applanix POSLV GNSS.}
  \label{fig:buick}
  \vspace*{-8mm}
\end{figure}

In this work, we present an unsupervised radar odometry pipeline that approaches the state of the art as reported on the Oxford Radar RobotCar Dataset \cite{barnes_oxford20}. Our network parameters are trained using only the on-board radar sensor, alleviating the need for an accurate groundtruth trajectory. To our knowledge, this is the first example of a totally unsupervised radar odometry pipeline. We show additional experimental results on 100 km of radar data collected on the data-taking platform shown in Figure~\ref{fig:buick}. We also provide a comparison of radar odometry performance in ideal and harsh weather conditions.\vspace{-0.5mm}

Our approach is based on the \ac{ESGVI} parameter learning framework of \citet{barfoot_ijrr20}, a nonlinear batch state estimation framework that provides a family of scalable estimators from a variational objective. Model parameters can be optimized jointly with the state using a data likelihood objective. \citet{yoon_ral21} recently applied the framework to train a deep network, demonstrating feature learning for lidar odometry with only on-board lidar data. We extend their methodology and apply it to radar odometry. Our approach is modular, enabling the use of modern deep learning tools and classical estimators with minimal interface requirements. Importantly, our method does not require the estimator to be differentiable, a limitation of some other learning-based methods \cite{barnes_corl19}.\vspace{-0.5mm}

The hybridization of deep learning and probabilistic state estimation allows us to achieve the best of both worlds. Deep learning can be leveraged to process rich sensor data, while classical estimators can be used to deal with probabilities and out-of-distribution samples through outlier rejection schemes, motion priors, and other estimation tools. Furthermore, classical estimators make the incorporation of additional sensors and constraints relatively straightforward.\vspace{-0.5mm}






We review related work in Section \ref{sec:related_work}. An overview of \ac{ESGVI} parameter learning is presented in Section \ref{sec:esgvi}, and the methodology of our radar odometry is in Section \ref{sec:method}. Experimental results are presented in Section \ref{sec:experiments}. Concluding remarks and future work are in Section \ref{sec:conclusion}.\vspace{-2mm}

\section{Related Work} \label{sec:related_work}

Initially, radar-based navigation was reliant on hand-tuned detectors specifically designed for radar \cite{clark_icra98, jose_iros04}. One such detector is CFAR (Constant False Alarm Rate), which uses a sliding window to find relative peaks in radar data \cite{rohling_irs11}. Although CFAR works well for detecting aircraft against a blank sky, it struggles to discriminate between clutter and objects of interest on the ground where the relative differences in radar cross section are low. These early methods relied primarily on location information to match detections and were thus limited by the accuracy of their data association.\vspace{-0.5mm}


Subsequent works have sought to improve feature detection and matching. \citet{checchin_fsr10} side-stepped this problem by applying correlative scan matching between entire radar scans via the Fourier-Mellin transform. Although correlation-based techniques have been shown to achieve excellent odometry performance, it remains challenging to apply them to large-scale mapping and localization due in part to their high data storage requirements during mapping.\vspace{-0.5mm}


Recent works have taken advantage of improvements to frequency-modulated continuous wave (FMCW) radar sensors. The radar sensor produced by Navtech \cite{navtech} is popular for navigation research due to its combination of high range resolution, angular resolution, range, and $360^\circ$ field of view.\vspace{-0.5mm}



\citet{cen_icra18} demonstrated large-scale outdoor ego-motion estimation with a scanning radar. Their work inspired a resurgence of research into radar-based perception and navigation systems. Research in this area has also been accelerated by the introduction of the Oxford Radar RobotCar Dataset \cite{barnes_oxford20}. Other datasets that feature a scanning radar include the MulRan Dataset \cite{kim_icra20} for multimodal place recognition, and the RADIATE Dataset \cite{sheeny_arxiv20} for object detection and tracking in adverse weather.\vspace{-0.5mm}

\citet{cen_icra19} later provided an update to their radar odometry pipeline with an improved gradient-based feature detector along with a new graph matching strategy. \citet{aldera_icra19} learn an attention policy to down-sample the number of measurements provided to the data association stage, thus speeding up the odometry pipeline. \citet{aldera_itsc19} train a classifier on the principal eigenvector of their graph matching problem in order to predict and correct for failures in radar odometry. \citet{park_icra20} applied the Fourier-Mellin Transform to log-polar images computed from down-sampled Cartesian images. \citet{barnes_corl19} demonstrated a fully differentiable, correlation-based radar odometry pipeline. Their approach learns a binary mask to remove distractor features before using brute force search to find the pose with the minimum cross correlation. \citet{hong_iros20} used vision-based features and graph matching to demonstrate the first radar-based SLAM system to operate in adverse weather. \citet{burnett_ral21} demonstrated that motion distortion and Doppler effects can have an adverse effect on radar-based navigation, but that these effects can be compensated using continuous-time estimation techniques.\vspace{-0.5mm}


These works represent significant steps towards radar-based autonomy. However, each of these methods relies on either hand-crafted feature detectors and descriptors or cumbersome phase correlation techniques. \citet{barnes_icra20} showed that learned features can result in superior radar odometry performance. Their work currently represents the state of the art for point-based radar odometry. Despite these results, their approach requires the estimator to be differentiable and they require groundtruth poses as a supervisory signal. Our approach does not suffer from either of these drawbacks.\vspace{-0.5mm}

Other works have focused on radar-based place recognition. \citet{saftescu_icra20} perform place recognition by learning rotationally invariant descriptors within a metric feature space. \citet{gadd_plans20} improve on this work by incorporating sequence-based place recognition based on SeqSLAM \cite{milford_icra12}. \citet{demartini_sens20} proposed a two-stage system to integrate topological localization with metric pose estimation. A related avenue of research has been to localize radar scans using existing satellite imagery \cite{tang_icra20} \cite{tang_rss20}. Further research has tackled radar-based perception through occupancy \cite{weston_icra19}, traversability \cite{williams_itsc20, broome_ai20}, and semantic segmentation \cite{kaul_arxiv20}.\vspace{-0.5mm}



\citet{barfoot_ijrr20} presented the \ac{ESGVI} framework and showed that model parameters can be jointly optimized along with the state using an \ac{EM} iterative optimization scheme on a data likelihood objective. In the E-step, model parameters are held fixed, and the state is optimized. In the M-step, the state distribution is held fixed, and the model parameters are optimized. \change{This idea originates from a line of work that applied \ac{EM} for linear system identification \citep{shumway1982approach,ghahramani96}}. \citet{ghahramani99} extended the idea to simple nonlinearities approximated with Gaussian radial basis functions. \citet{wong_ral20b} first applied \ac{EM} under the \ac{ESGVI} framework to large-scale trajectory estimation by learning noise models robust to outliers. \citet{yoon_ral21} demonstrated the first application of \ac{ESGVI} parameter learning to a deep network, where lidar features for odometry were learned without groundtruth. We apply their methodology and adapt it for radar odometry.\vspace{-0.5mm}

Our framework shares similarities with the \ac{VAE} \cite{kingma2013auto} framework, as both start from a data likelihood objective and optimize the \ac{ELBO}. In contrast to the \ac{VAE}, which approximates latent state inference with a network, our framework applies classic state estimation (e.g., factor graph optimization). Extensions to the \ac{VAE} for problems with graphical structure exist \cite{johnson2016composing}, but our method makes use of existing estimation tools familiar to the robotics field. Most similar to our framework is the work of \citet{detone2018b}. They present a self-supervised visual odometry framework that has a deep network frontend trained according to a bundle adjustment backend. Compared to their framework, ours is based on a probabilistic objective and can \change{handle} uncertainty in the posterior estimates.\vspace{-0.5mm}




\section{Exactly Sparse Gaussian Variational Inference Parameter Learning} \label{sec:esgvi}

This section summarizes parameter learning in the \ac{ESGVI} framework as presented in prior work \cite{barfoot_ijrr20,wong_ral20b,yoon_ral21}. The loss function is the negative log-likelihood of the observed data,
\begin{equation} \label{eq:data_likelihood}
	\mathscr{L} = -\ln{p(\mbf{z} | \mbs{\theta})},  
\end{equation}
where $\mbf{z}$ is the data \change{(e.g., radar measurements)}, and $\mbs{\theta}$ are the parameters \change{(e.g., network parameters)}. We apply the usual \ac{EM}\footnote{While the acronym stays the same, we work with the negative log-likelihood and are technically applying Expectation Minimization.} decomposition after introducing the latent trajectory, $\mbf{x}$, which is written as
\begin{equation}
	\mathscr{L} = \begin{medsize}
		\underbrace{\int^{\mbs{\infty}}_{-\mbs{\infty}} q(\mbf{x}) \ln \left( \frac{p(\mbf{x} | \mbf{z}, \mbs{\theta})}{q(\mbf{x})} \right) d\mbf{x}}_{\mbox{$\leq$ 0}}
		\underbrace{-\int^{\mbs{\infty}}_{-\mbs{\infty}} q(\mbf{x}) \ln \left( \frac{p(\mbf{x}, \mbf{z}  | \mbs{\theta})}{q(\mbf{x})} \right) d\mbf{x}}_{\mbox{upper bound}}, \label{eq:em_decomp} 
	\end{medsize}
\end{equation}
where we define our posterior approximation as a multivariate Gaussian distribution, $q(\mbf{x}) = \mathcal{N}(\mbs{\mu}, \mbs{\Sigma})$.

We work with the upper bound term, commonly referred to as the (negative) \ac{ELBO}. Applying the definition of entropy of a Gaussian and dropping constants, we can rewrite the term as the \ac{ESGVI} loss functional,
\begin{equation}
	\label{eq:functional}
	V(q|\mbs{\theta}) = \mathbb{E}_q[ \phi(\mbf{x}, \mbf{z}|\mbs{\theta})] + \frac{1}{2} \ln \left( |\mbs{\Sigma}^{-1}| \right),
\end{equation}
where $\mathbb{E}[\cdot]$ is the expectation operator, $|\cdot|$ is the matrix determinant, and we define the joint factor $\phi(\mbf{x}, \mbf{z}|\mbs{\theta}) = - \ln p(\mbf{x},\mbf{z}|\mbs{\theta})$.

We apply \ac{GEM} \cite{neal98}, a variation of \ac{EM} that does not run the M-step to completion (convergence), to gradually optimize the data likelihood, $\mathscr{L}$. In the E-step, we hold $\mbs{\theta}$ fixed and optimize $V(q|\mbs{\theta})$ for $q(\mbf{x})$. In the M-step, we hold $q(\mbf{x})$ fixed and optimize $V(q|\mbs{\theta})$ for $\mbs{\theta}$. When the expectation over the posterior, $q(\mbf{x})$, is approximated at the mean of the Gaussian, the E-step is the familiar \ac{MAP} estimator \cite{barfoot_ijrr20}.

\begin{figure} [t]
	\centering
	
	\begin{tikzpicture} [connector/.style={>=latex, line width=1.4pt},
		squarenode/.style={draw, ultra thick, rectangle, minimum size=8mm},
		roundnode/.style={draw, ultra thick, circle, minimum size=11mm},
		starnode/.style={draw, thick, star, star points = 5, star point height=1.5mm, minimum size=1.5mm},
		dot/.style={circle, fill, minimum size=#1, inner sep=0pt, outer sep=0pt},
		trapnode/.style={draw=none, trapezium, trapezium angle=60, fill=blue!60, minimum width=8mm, align=center}
		]
		
		\node[roundnode, fill=black!20] (x1) {$\mathbf{x}_{k}$};
		\node[roundnode, fill=white] (x2) [right=of x1] {$\mathbf{x}_{k+1}$};
		\node[roundnode, fill=white] (x3) [right=of x2] {$\mathbf{x}_{k+2}$};
		\draw[-, connector] (x1.east) -- (x2.west) node [midway] (x12) {} node [midway, below] {$\phi^p$};
		\draw[-, connector] (x2.east) -- (x3.west) node [midway] (x23) {} node [midway, below] {$\phi^p$};
		\node[dot=5pt] at (x12) {};
		\node[dot=5pt] at (x23) {};
		
		\draw[-, connector, bend left=45] (x1.north) to node [midway] (m11) {} node [midway, below] {$\phi^m$} (x2.north);
		\node[dot=5pt] at (m11) {};
		\draw[-, connector, bend left=45] (x2.north) to node [midway] (m22) {} node [midway, below] {$\phi^m$} (x3.north);
		\node[dot=5pt] at (m22) {};
		
		\node[starnode] (star1) [above=of x1, yshift=4mm] {};
		\node[starnode] (star2) [above=of x2, yshift=4mm] {};
		\node[starnode] (star3) [above=of x3, yshift=4mm] {};
		
		\draw[->, connector, bend left = 45] (star1.south east) to (m11);
		\draw[->, connector, bend right = 45] (star2.south west) to (m11);
		\draw[->, connector, bend left = 45] (star2.south east) to (m22);
		\draw[->, connector, bend right = 45] (star3.south west) to (m22);
		
		\node[trapnode, shape border rotate=180] (dnn1a) [above=of star1, yshift=-2mm] {};
		\node[trapnode] (dnn1) [above=of star1, yshift=-5mm] {};
		\node (t1) at ($(dnn1)!0.5!(dnn1a) + (0.45, 0)$) {$\mbs{\theta}$};
		\node[trapnode, shape border rotate=180] (dnn2a) [above=of star2, yshift=-2mm] {};
		\node[trapnode] (dnn2) [above=of star2, yshift=-5mm] {};
		\node (t2) at ($(dnn2)!0.5!(dnn2a) + (0.45, 0)$) {$\mbs{\theta}$};
		\node[trapnode, shape border rotate=180] (dnn3a) [above=of star3, yshift=-2mm] {};
		\node[trapnode] (dnn3) [above=of star3, yshift=-5mm] {};
		\node (t3) at ($(dnn3)!0.5!(dnn3a) + (0.45, 0)$) {$\mbs{\theta}$};
		
		\node[inner sep=0pt] (data1) [above of=dnn1a, yshift=8mm] {\includegraphics[width=.22\columnwidth]{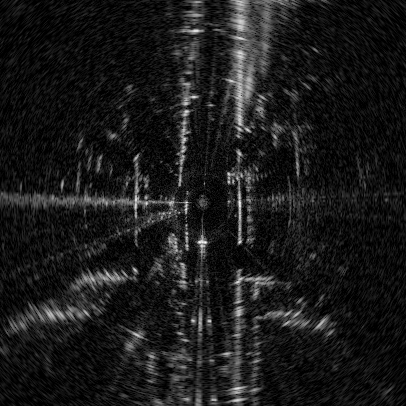}};
		\node[inner sep=0pt] (data2) [above of=dnn2a, yshift=8mm] {\includegraphics[width=.22\columnwidth]{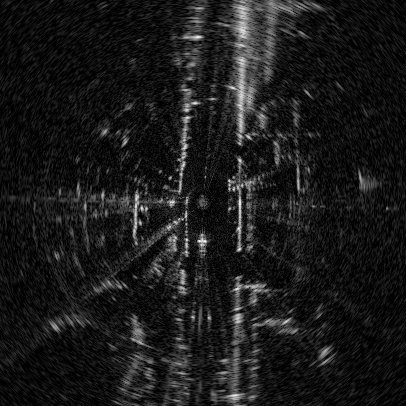}};
		\node[inner sep=0pt] (data3) [above of=dnn3a, yshift=8mm] {\includegraphics[width=.22\columnwidth]{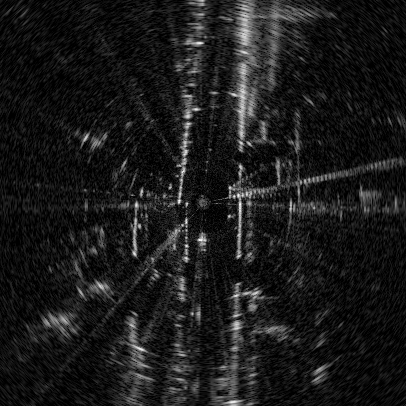}};
		\node[above of=data1, yshift=2mm] {$\mathbf{z}_k$};
		\node[above of=data2, yshift=2mm] {$\mathbf{z}_{k+1}$};
		\node[above of=data3, yshift=2mm] {$\mathbf{z}_{k+2}$};
		
		\node (bp1) [right of=x3] {};
		\node (bp2) [right of=dnn3a] {};
		\draw[->, line width=1.6pt, color=blue!60, text=black] (bp1) -- (bp2) node [midway, right, text width=15mm] {Unsupervised Backprop. (M-Step)};
		
		\draw[->, connector] (dnn1.south) -- (star1.north);
		\draw[->, connector] (dnn2.south) -- (star2.north);
		\draw[->, connector] (dnn3.south) -- (star3.north);
		\draw[->, connector] (data1.south) -- (dnn1a.north);
		\draw[->, connector] (data2.south) -- (dnn2a.north);
		\draw[->, connector] (data3.south) -- (dnn3a.north);
		
		\node (text1) [below of=x2, align=center, text width=45mm] {Estimated Trajectory (E-Step)};
		
	\end{tikzpicture}
	\caption{This figure depicts the factor graph of our radar odometry pipeline. $\mbf{x}_k$ and $\mbf{z}_k$ are defined as the state of the vehicle and the radar scan at time $t_k$, respectively. The vehicle trajectory is estimated over a sliding window of $w$ frames, where $w = 3$ in this figure. The deep network parameters are denoted by $\mbs{\theta}$. The output of the network is a set of feature locations $(x_i, y_i)$, their associated inverse covariance matrices, $\mbf{W}_i$, and their learned descriptors, $\mbf{d}_i$, which are together represented by stars in the diagram. These features are then matched between pairs of frames using a differentiable softmax matcher. The matched features are then used to form measurement factors, $\phi^m$. A white-noise-on-acceleration motion prior is applied to create prior factors, $\phi^p$.}
	\label{fig:pose_graph}
\end{figure}
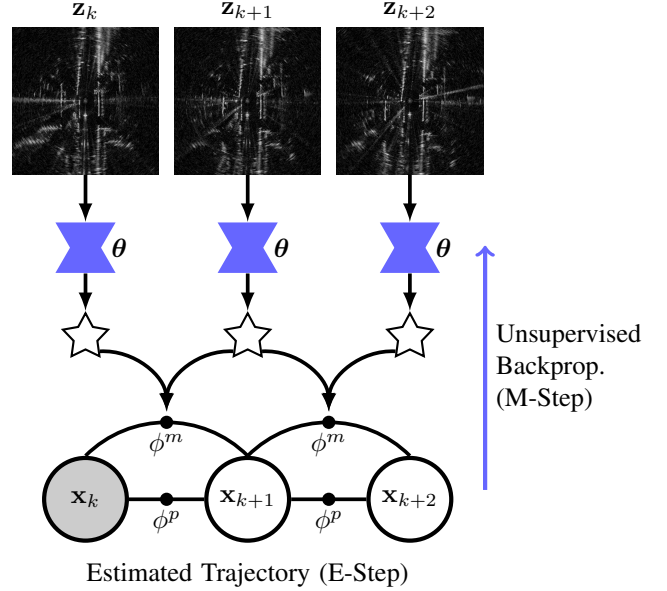

\section{Unsupervised Deep Learning for Radar Odometry} \label{sec:method}

\subsection{Problem Definition} \label{subsec:problem}
In this subsection, we summarize the sliding-window odometry formulation presented by \citet{yoon_ral21}. The state we estimate at time $t_k$ is $\mbf{x}_k = \{\mbf{T}_{k,0}, \mbs{\varpi}_k\}$, where the pose $\mbf{T}_{k,0} \in SE(3)$ is a transformation between frames at $t_k$ and $t_0$, and $\mbs{\varpi}_k \in \mathbb{R}^6$ is the body-centric velocity\footnote{Despite the radar being a 2D sensor, we formulate our problem in $SE(3)$ to be compatible with other 3D sensor modalities.}. We optimize a sliding-window of $w$ frames, $t_\tau, \dots, t_{\tau+w-1}$, where each state has a corresponding radar scan. The first pose of the window, $\mbf{T}_{\tau,0}$, is locked (not optimized) and treated as the reference frame for keypoint matching. 

Our joint factor, $\phi(\mbf{x}, \mbf{z}|\mbs{\theta})$, splits into motion prior factors and measurement factors. Figure~\ref{fig:pose_graph} shows an example factor graph illustration. We write $\phi(\mbf{x}, \mbf{z}|\mbs{\theta})$ as
\begin{equation}
  \sum_{k=\tau+1}^{\tau+w-1} \left( \phi^p(\mbf{x}_{k-1}, \mbf{x}_k) + \sum_{\ell=1}^{L_k} \phi^m(\mbf{z}_k^\ell, \mbf{r}_{\tau}^\ell | \mbf{x}_\tau, \mbf{x}_k,  \mbs{\theta}) \right),
\end{equation}
where $\mbf{z}_k^\ell$ is the $\ell$th keypoint measurement in frame $k$, which has a total of $L_k$ keypoints, and $\mbf{r}_{\tau}^\ell$ is its matched point from frame $\tau$.

For the motion prior factors, $\phi^p$, we apply a white-noise-on-acceleration prior as presented by Anderson and Barfoot \cite{anderson_iros15}, which is defined by the following kinematic equations:
\begin{equation} \label{eq:se3_prior}
	\begin{split}
		\dot{\mathbf{T}}(t)&=\mbs{\varpi}(t)^\wedge{}\mathbf{T}(t), \\
		\dot{\mbs{\varpi}}&=\mathbf{w}(t),\quad \mathbf{w}(t) \sim \mathcal{GP}(\mathbf{0}, \mathbf{Q}_c\delta(t-t')),
	\end{split}
\end{equation}
where $\mathbf{T}(t_k) = \mathbf{T}_{k,0}$, and \(\mbf{w}(t) \in \mathbb{R}^6\) is a zero-mean, white-noise Gaussian process. The operator, $\wedge$, transforms an element of $\mathbb{R}^6$ into a member of Lie algebra, $\mathfrak{se}(3)$ \cite{Barfoot2017}:
\begin{equation}
  \bbm \mbf{u} \\ \mbf{v} \ebm^{\wedge} = \bbm \mbf{v}^{\wedge} & \mbf{u} \\ \mbf{0}^T & 0 \ebm, \quad \mbf{v}^{\wedge} = \begin{medsize} 
    \bbm v_1 \\ v_2 \\ v_3 \ebm^{\wedge} = \bbm 0 & -v_3 & v_2 \\ v_3 & 0 & -v_1 \\ -v_2 & v_1 & 0 \ebm 
  \end{medsize},
\end{equation}
where $\wedge$ is also overloaded as the skew-symmetric operator for elements of $\mathbb{R}^3$.

The measurement factors, $\phi^m$, are of the form:
\begin{align}
 \label{eq:meas_factor}
  &\phi^m(\mbf{z}_k^\ell, \mbf{r}_{\tau}^\ell | \mbf{x}_\tau, \mbf{x}_k, \mbs{\theta}) = \frac{1}{2} 
  {\mbf{e}_k^\ell}^T \mbf{W}_k^\ell \mbf{e}_k^\ell - \ln\left| \mbf{W}_k^\ell \right|, \\ \label{eq:meas_error}
  &\mbf{e}_k^\ell = \mbf{D} \left( \mbf{z}_k^\ell - \mbf{T}_{k,0} \mbf{T}_{0,\tau} \mbf{r}_{\tau}^{\ell_k} \right),
\end{align}
where we use the log-likelihood of a Gaussian as the factor, and $\mbf{D}$ is a $3 \times 4$ constant projection matrix that removes the homogeneous element. The homogenous keypoint, $\mbf{z}_k^\ell$, its point match, $\mbf{r}_{\tau}^{\ell_k}$, and its inverse covariance \change{(weight)} matrix, $\mbf{W}_k^\ell$, are quantities that depend on the network parameters, $\mbs{\theta}$.

\begin{figure*}
  \centering
  \includegraphics[width=0.98\textwidth]{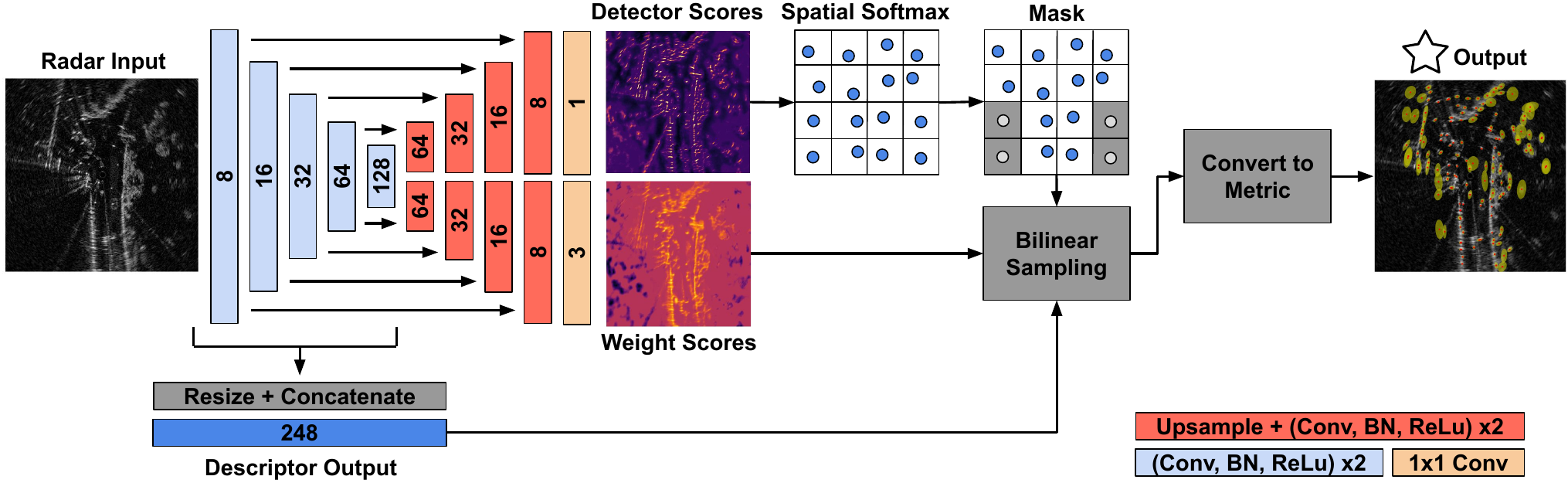}
  \caption{We based our network on the architecture presented by \citet{barnes_icra20}. The network outputs detector scores for keypoint detection, weight scores predicting keypoint uncertainty, and descriptors for matching. The weight scores are composed into $2\times2$ inverse covariance matrices (see (\ref{eq:LDL}), the corresponding image is the log-determinant). Descriptors are the concatenation of all encoder layer outputs after resizing via bilinear interpolation. The encoder and decoder layers are a double application of a $3\times3$ convolution, batch normalization, and ReLU nonlinearity. The layer sizes vary by a factor of 2 through max-pooling (encoder) and bilinear upsampling (decoder). Note that the output size is the same as the input, and are visually smaller in the interest of space. An output $1\times1$ convolution is applied for the detector and weight scores. The detector score map is partitioned into uniform cells, where a spatial softmax and weighted summation of coordinates are applied to yield a keypoint for each cell. Corresponding weights and descriptors are obtained via bilinear sampling.}
  \label{fig:network}
\end{figure*}

\subsection{Network} \label{subsec:network} \vspace{-1mm}
Our network is based on the architecture presented by \citet{barnes_icra20}, which is a U-Net \cite{unet} style convolutional encoder-multi-decoder architecture to output radar keypoints, weights, and descriptors. The input to the network is a 2D Cartesian projection of the polar radar data. Example radar scans are shown at the top of Figure \ref{fig:pose_graph}.\vspace{-0.5mm}

An illustration of our network architecture is shown in Figure \ref{fig:network}. A dense descriptor map is created by resizing the output of each encoder block before concatenation into a 248-channel tensor. In our approach, the weight score is a 3-channel tensor and the detector score is a 1-channel tensor. The detector score tensor is then partitioned into $(N = 400)$ equally sized cells, with each producing a candidate 2D keypoint. A spatial softmax within each cell, followed by a weighted summation of pixel coordinates produces the image-space keypoint coordinates of each cell. The corresponding descriptor and weight score vectors are bilinearly sampled using the keypoint coordinates. The image coordinates are converted to metric coordinates using the known m/pixel resolution. Finally, we formulate the 3D homogenous keypoint, $\mbf{z}_k^\ell$, by appending a $0$ as a third coordinate, and a $1$ as the fourth homogenous element.\vspace{-0.5mm}

Our detector produces a candidate keypoint for each square cell in a uniformly partitioned radar image. This leads to candidate keypoints in regions of the image that are void of data (i.e., black regions of a radar image). As we are training without groundtruth, we found it necessary in practice to mask (reject) these keypoints. We threshold each azimuth of the polar radar scan by a scalar multiple, $\beta = 3$, of its mean intensity, where exceeding the threshold is considered valid. Projecting the result into Cartesian space produces a binary mask of valid pixels. We then threshold on the ratio of valid pixels in each square cell for robustness to noise. Keypoints belonging to cells with less than 5\% valid pixels are rejected.

\begin{figure}[t]
	\centering
	\includegraphics[width=0.47\columnwidth, trim=2cm 2cm 2cm 2cm, clip]{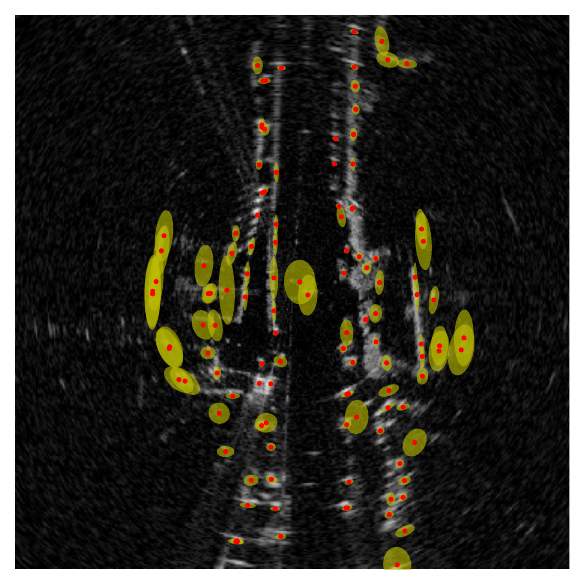}
	\includegraphics[width=0.47\columnwidth, trim=2cm 2cm 2cm 2cm, clip]{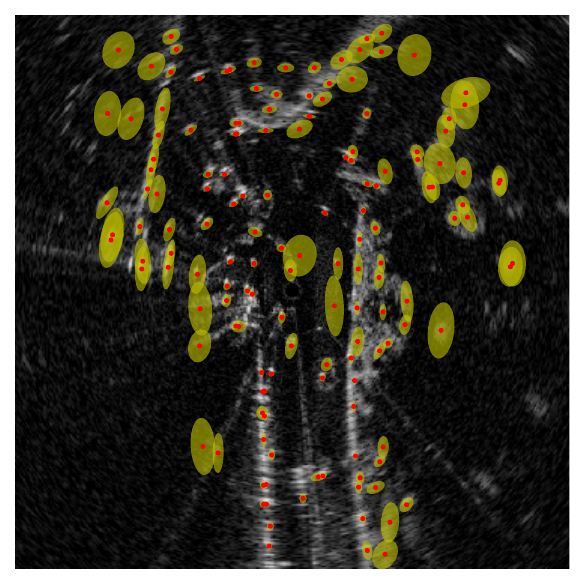} \\
	\vspace*{0.1cm}
	
	\centering
	\includegraphics[width=0.47\columnwidth, trim=2cm 2cm 2cm 2cm, clip]{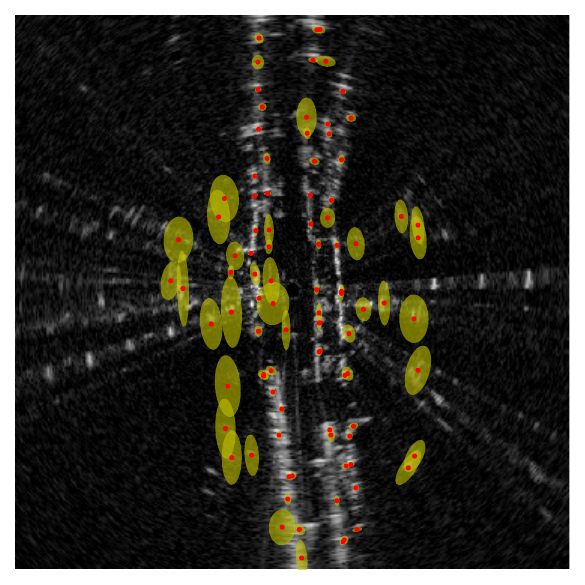}
	\includegraphics[width=0.47\columnwidth, trim=2cm 2cm 2cm 2cm, clip]{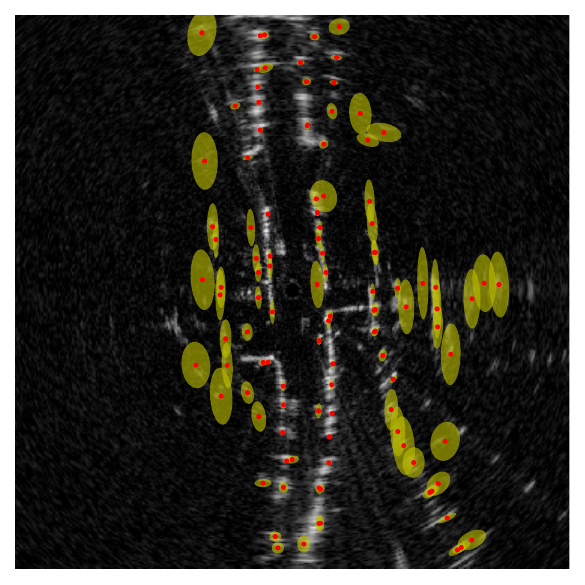}
	\caption{A visualization of keypoints (red) on radar images with 5 standard deviation uncertainty ellipses (yellow). Many keypoints have elongated uncertainties consistent to the scene geometry, and as expected, keypoints further away from the sensor (image center) are more uncertain.} 
	\label{fig:ellipse}
	\vspace{-8mm}
\end{figure}

A dense match between each keypoint descriptor and the reference descriptor tensor is applied with a softmax to preserve differentiability. We compute the dot product between each keypoint descriptor and all descriptors of the reference:
\begin{equation}
  \mbf{c}_k^{\ell^T} = \mbf{d}^{\ell \, T}_k \bbm \mbf{d}_{\tau}^1 & \cdots & \mbf{d}_{\tau}^N \ebm,
\end{equation}
where $\mbf{d}^\ell_k$ is the descriptor vector of keypoint $\mbf{z}^\ell_k$, and $\mbf{d}_{\tau}^1, \dots, \mbf{d}_{\tau}^N$ are the descriptor vectors of the reference. We apply a softmax on $\mbf{c}_k^\ell$ and compute a weighted summation. The reference match for keypoint $\mbf{z}^\ell_k$ is therefore
\begin{equation}
  \mbf{r}_{\tau}^{\ell_k} = \bbm \mbf{p}^1_\tau & \cdots & \mbf{p}^N_\tau \ebm \times~\mbox{softmax}(T \mbf{c}_{k}^\ell),
\end{equation}
where $T = 100$ is a softmax temperature constant, $\mbf{p}^1_\tau, \dots, \mbf{p}^N_\tau$ are the homogeneous reference coordinates, and $\mbf{p}^n_\tau \in \mathbb{R}^4$.
\vspace{1mm}
The weight score vectors of each keypoint are assembled into matrices with the following decomposition \cite{liu2018deep,yoon_ral21}:
\vspace{1mm}
\begin{equation} \label{eq:LDL}
\mbf{W}_k^\ell = \begin{medsize} \bbm \mbf{R} & \mbf{0} \\ \mbf{0}^T & c \ebm \end{medsize} , \quad
\mbf{R} = \begin{medsize}
  \bbm 1 & 0 \\ d_3 & 1 \ebm
  \bbm \exp d_1 & 0 \\ 0 & \exp d_2 \ebm
  \bbm 1 & 0 \\ d_3 & 1 \ebm^T, 
\end{medsize} 
\end{equation}
\vspace{1mm}where $\left( d_1, d_2, d_3 \right)$ is the weight score vector corresponding to keypoint $\mbf{z}_k^\ell$, and $c = 10^4$ is a constant corresponding to the (inverse) variance of the third coordinate (which will always be 0). Visualizations of the learned keypoint covariances, $\mbf{R}^{-1}$, are shown in Figure \ref{fig:ellipse} as uncertainty ellipses.

\vspace{1mm}

\subsection{Training and Inference}
We follow the training methodology of \citet{yoon_ral21}, which blends the \ac{GEM} iterative optimization scheme required by \ac{ESGVI} with \ac{SGD} applied in conventional network training. The E-step, which optimizes for the current best posterior estimate, $q(\mbf{x})$, is simply included as part of the forward propagation routine.

\newpage
Windows of radar scans are randomly sampled as a mini-batch of data. Forward propagation is summarized as:
\vspace{-0.5mm}
\begin{enumerate}
\item Traditional forward propagation of the network to output radar features (see Section \ref{subsec:network}).
\vspace{-0.5mm}
\item Construct the motion prior factors, $\phi^p$, and measurement factors, $\phi^m$,  (see Section \ref{subsec:problem} and \ref{subsec:network}).
\vspace{-0.5mm}
\item Batch inference for the current best posterior estimate $q(\mbf{x})$ for each window (E-step).
\vspace{-0.5mm}
\end{enumerate}
\vspace{-0.5mm}
We approximate the E-step with the Gauss-Newton algorithm, which involves approximations to the Hessian and approximating the expectation in (\ref{eq:functional}) at only the mean of the posterior.


The M-step is network backpropagation on the loss functional (\ref{eq:functional}), which only applies to the measurement factors since the motion prior factors are constant with respect to the parameters, $\mbs{\theta}$. Similar to the E-step, we approximate the expectation at only the mean of the posterior. Intuitively, we are using our best-guess trajectory as a supervisory signal to then carry out standard \ac{SGD} to train the feature network.
\vspace{-3mm}

\subsection{Outlier Rejection} \label{subsec:outlier}
\vspace{-1mm}

We apply the Geman-McClure robust cost function on the measurement factors, $\phi^m$, in the E-step for outlier rejection. For the M-step, we follow \citet{yoon_ral21} and apply a constant threshold, $\alpha$, on the squared Mahalanobis distance of the measurement factor with the current best posterior estimate,
\vspace{-1.5mm}
\begin{equation} \label{eq:mah_thres}
{\mbf{e}_k^\ell}^T \mbf{W}_k^\ell \mbf{e}_k^\ell  > \alpha,
\end{equation}
\vspace{-1.0mm}where $\mbf{e}_k^\ell$ is as defined in (\ref{eq:meas_error}). We do not backpropagate keypoint matches that are greater than $\alpha = 16$.

After training, we improved odometry performance by rejecting keypoints with inverse covariances that have a log determinant, $\text{log}|\mbf{R}|$, less than a threshold $(\eta = 4.0)$. \change{We also use RANSAC at test time and only estimate on the inliers.}


\section{Experimental Results} \label{sec:experiments}

\subsection{Experiment Setup}
We evaluate our approach, \ac{HERO}, on the publicly available Oxford Radar RobotCar Dataset \cite{barnes_oxford20} and on \change{our own dataset, collected using the} data-taking platform in Figure~\ref{fig:buick}. The Oxford dataset is divided into 32 sequences, each approximately 10 km in length. We follow existing work \cite{barnes_icra20} by training on 24 sequences, validating on 1 sequence, and testing on the same 7 sequences as \cite{barnes_icra20}.

Our implementation is a hybrid between Python and C++, where network-related code is implemented in Python using PyTorch \cite{pytorch}, and estimation-related code is implemented using STEAM\footnote{\url{https://github.com/utiasASRL/steam}}, an open-source C++ estimation library. When training the network, we use a fixed random seed and the Adam optimizer \cite{Kingma2014} with a learning rate of $1 \times 10^{-5}$ and a mini-batch size of 1 window ($w = 4$) for up to 100k iterations. The Navtech radar sensor used in the Oxford dataset has a range resolution of $0.0438$ m/bin. To minimize projection errors, we chose a Cartesian resolution of $0.2628$ m/pixel to be an integer multiple of the radar resolution. The Cartesian radar images are made to be square with a width of 640 pixels. For the spatial softmax operation, each cell is $32 \times 32$ pixels, resulting in 400 total keypoints. We use a softmax temperature of 100. \change{For the motion prior, $\mathbf{Q}_c$ is made to be diagonal. The entries of $\mathbf{Q}_c^{-1}$ can be thought of as penalty terms on body-centric linear and angular acceleration. It is possible to learn $\mathbf{Q}_c$ as is done \cite{wong_ral20b}. However, in our implementation, the values are hand-tuned.} Similar to \citet{barnes_icra20}, we augment our training data with random rotations of up to 0.26 radians. Our sliding-window implementation\footnote{On an Nvidia Tesla V100 GPU and 2.2 GHz Intel Xeon CPU.} takes on average 0.07 seconds, 0.13 seconds, and 0.18 seconds for window sizes of 2, 3, and 4, respectively. Since the radar sensor spins at 4 Hz, our current implementation is real-time capable. During evaluation, we use the timestamp of each measurement to do continuous-time estimation with STEAM to compensate for motion distortion. For more information on this approach, we refer readers to the work of Sean Anderson et al. \cite{anderson_iros15} \cite{anderson_robots15}.

\vspace{-2mm}


\renewcommand{\arraystretch}{1.2}
\begin{table}[t]
	\centering
	\caption{Radar Odometry Results. (HC): Hand-crafted, (L): Learned. \vspace{-3mm}}
	\begin{tabular}{|c|c|c|c|}
		\hline
		\textbf{Methods} & \textbf{Supervision}          & \multicolumn{1}{c|}{\begin{tabular}[c]{@{}c@{}}\textbf{Trans.}\\ \textbf{Error (\%)}\end{tabular}} & \multicolumn{1}{c|}{\begin{tabular}[c]{@{}c@{}}\textbf{Rot. Error}\\ \textbf{(deg/1000m)}\end{tabular}} \\ \hline
		UnderTheRadar \cite{barnes_icra20}	& Supervised (L)		& 2.0583                & 6.7				\\ 
		Cen RO \cite{cen_icra18}         	 	& Unsupervised (HC) 		& 3.7168                & 9.5				\\ 
		MC-RANSAC \cite{burnett_ral21}       	& Unsupervised (HC) 		& 3.3190                & 10.93				\\
		CFEAR \cite{adolfsson_arxiv21}       	& Unsupervised (HC) 		& \textbf{1.76}                & \textbf{5.0}				\\
		HERO (Ours)            					& Unsupervised (L) 		& 1.9879		& 6.524	\\ \hline
	\end{tabular}
	\label{tab:odom}\vspace{-6mm}
\end{table}

\subsection{Oxford Radar RobotCar Dataset}
\begin{figure}[t]
	\includegraphics[width=0.98\columnwidth]{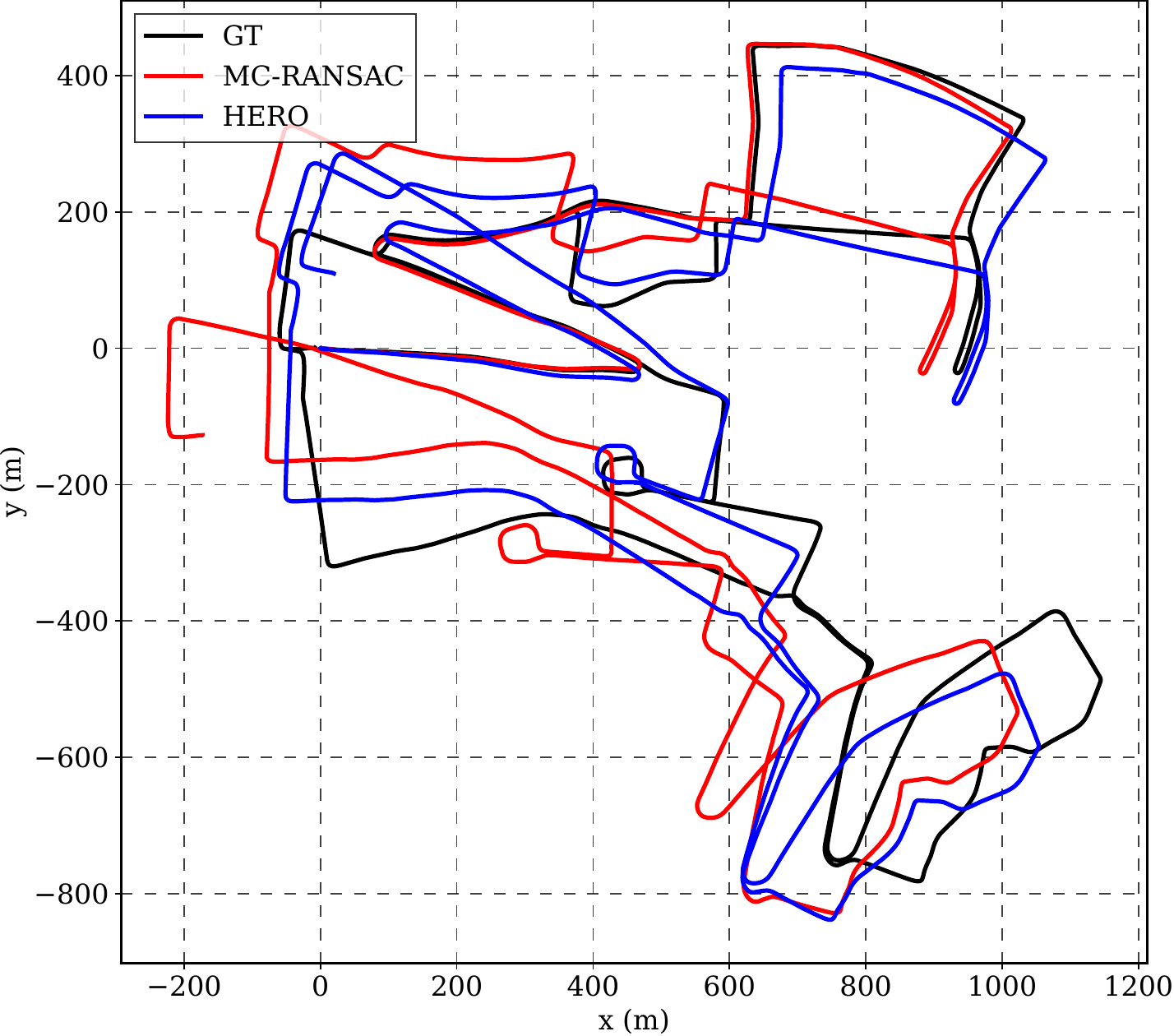}
	\centering
	\caption{This figure depicts the results of our unsupervised Hybrid-Estimate Radar Odometry (HERO) in blue alongside the results of a hand-crafted radar odometry pipeline based on motion compensated RANSAC (MC-RANSAC) \cite{burnett_ral21} in red. The groundtruth trajectory is plotted in black (GT). Sequence: 2019-01-10-14-02-34-radar-oxford-10k.}
	\label{fig:odom}
	\vspace{-6.5mm}
\end{figure}

\vspace{-1.5mm}
Here, we compare our radar odometry results with other point-based radar odometry methods including \citet{cen_icra18} (Cen~RO), \citet{barnes_icra20} (Under the Radar), CFEAR \cite{adolfsson_arxiv21}, and \citet{burnett_ral21} (MC-RANSAC). Following these previous works, we report our results using the KITTI odometry metrics \cite{Geiger2012}, which average the relative position and orientation errors over every sub-sequence of length ($100~\text{m}, 200~\text{m},~\cdots~, 800~\text{m}$). The results in Table~\ref{tab:odom} show that our method, \ac{HERO}, is competitive with other unsupervised radar odometry methods, surpassing the hand-crafted algorithms Cen~RO and MC-RANSAC. In addition, our method exceeds the performance of Under the Radar, a learned point-based radar odometry approach, without needing any groundtruth supervision. This feature gives our method an advantage for deployment in regions where a source of high-quality groundtruth is unavailable. Sources of groundtruth, such as a GPS/INS system, are not only costly, but require a clear line of sight to the sky for GPS reception. Notably, the authors of the Oxford Radar RobotCar Dataset \cite{barnes_icra20} state that the accuracy of their GPS/INS system varied significantly due to poor GPS reception \cite{Maddern2017}. They addressed this issue by including visual odometry and loop closures into a large scale optimization \cite{barnes_icra20}. Figure \ref{fig:odom} illustrates an example sequence comparing our odometry method to MC-RANSAC.


We provide an ablation study of our method in Table~\ref{tab:ablation}. We first compare our baseline to the following variations:
\vspace{-0.5mm}
\begin{itemize}
  \item[$\bullet$] \textbf{Scalar Weight}: Instead of learning a $2\times2$ weight (inverse covariance) matrix (see (\ref{eq:LDL})), we learn a scalar weight.
  \item[$\bullet$] \textbf{No Mah. Threshold}: We do not apply the outlier rejection threshold on the squared Mahalanobis distance \ref{eq:mah_thres}.
  \item[$\bullet$] \textbf{No Masking}: We do not mask keypoints in void regions of the radar images, as described in Section \ref{subsec:network}.
  \item[$\bullet$] \textbf{No Augmentation}: We do not augment our training data with random rotations.
\end{itemize}
We also show the effect of varying the window size of our sliding-window optimization and the resolution of the projected radar images. Our baseline uses a window size of 4 and a resolution of 0.2628 m/pixel.
\vspace{-0.5mm}

The results in Table~\ref{tab:ablation} suggest that rotation augmentation, masking keypoints, and thresholding on Mahalanobis distance were the most significant hyperparameters. We theorize that because a large fraction of the Oxford dataset consists of either driving straight or waiting at a red light, training without rotation augmentation makes it less likely that the network will learn to properly match features across large frame-to-frame rotations. Putting a hard threshold on the Mahalanobis distance before backpropagation allows us to only backpropagate the errors from likely inliers. The masking of keypoints from empty regions of the input radar scan was necessary to achieve good odometry results. Our interpretation is that adding this additional structure to the learning problem is needed for the unsupervised model to succeed.


\vspace{-2mm}

\renewcommand{\arraystretch}{1.1}
\begin{table}[t]
	\centering
	\caption{Ablation study. Results for sequence 2019-01-10-14-02-34-radar-oxford-10k.\vspace{-2mm}}
	\begin{tabular}{|l|c|c|}
		\hline
		\textbf{Configuration}             & {\begin{tabular}[c]{@{}c@{}}\textbf{Translational}\\ \textbf{Error (\%)}\end{tabular}} & \multicolumn{1}{c|}{\begin{tabular}[c]{@{}c@{}}\textbf{Rotational Error}\\ \textbf{(deg/1000m)}\end{tabular}} \\ \hline
		Baseline           & 2.262                  & 7.601                    \\
		Scalar Weight      & 2.484                  & 7.997                    \\
		No Mah. Threshold  & 3.083                  & 9.300                    \\
		No Masking         & 3.203                  & 10.07                    \\
		No Augmentation    & 5.054                  & 16.60                    \\
		Window Size 2           & 2.488                  & 7.901                    \\
		Window Size 3           & 2.393                  & 7.657                    \\
		Cart Res: 0.2160             & 2.396                  & 7.602                    \\
		Cart Res: 0.3024             & 2.407                  & 7.927                    \\
		Baseline + $(\text{log}|\mbf{R}| < 4.0)$ & 1.953 & 6.532 \\ \hline
	\end{tabular}
	\label{tab:ablation}
	\vspace{-7mm}
\end{table}

\subsection{Additional Experiments on the Boreas Dataset}
In this section, we provide additional experimental results for our radar odometry using 100 km of driving obtained in an urban environment\footnote{We plan on making our dataset publicly available in the next year.}. The dataset was collected using the platform shown in Figure~\ref{fig:buick}, which includes a Velodyne Alpha-Prime (128-beam) lidar, Navtech CIR204-H radar, FLIR Blackfly~S monocular camera, and Applanix POSLV GNSS. This Navtech sensor has a range resolution of 0.0596~m and a total range of 200~m, compared to the 0.0438~m resolution and 163~m range of the Navtech sensor used in the Oxford dataset. We do not require groundtruth position data to train our network, but obtain it for the test sequences in order to calculate KITTI odometry drift metrics. Groundtruth positioning is obtained from the post-processed GNSS results and is accurate to within 12~cm. 

\vspace{-0.5mm}
The 100 km of total driving data is divided into 11 individual sequences, each approximately 9 km in length. We use 7 sequences for training, 1 for validation, and 3 for testing. Of the 3 test sequences, two were taken during a snow storm, and the other was taken on a sunny day.
\vspace{-0.5mm}

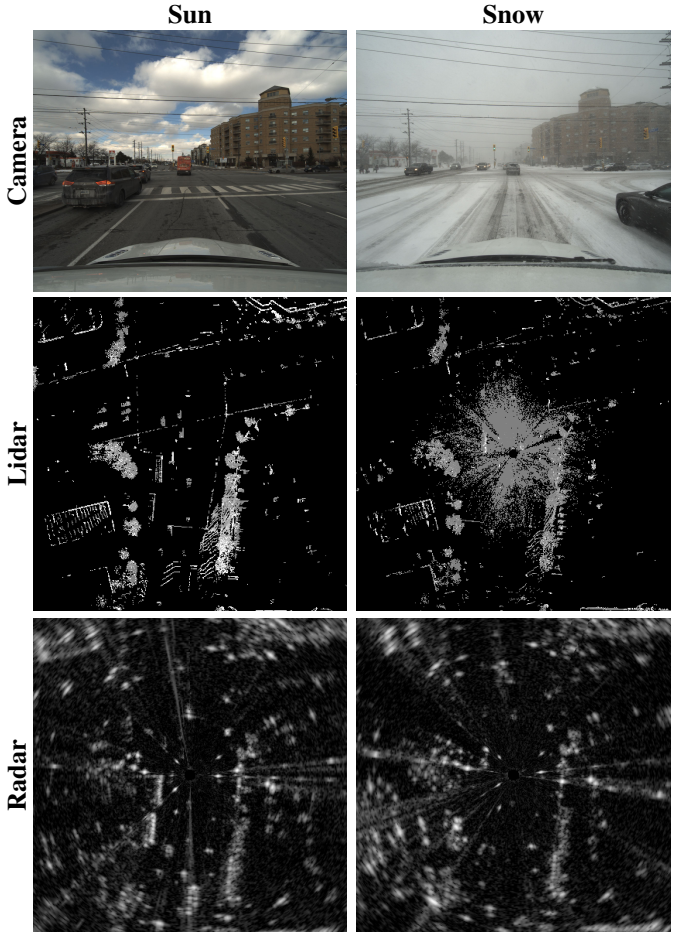
\begin{figure} [t]
	\centering
	\begin{tikzpicture}

		\node[inner sep=0pt] (cam1)
			{\includegraphics[width=0.47\columnwidth]{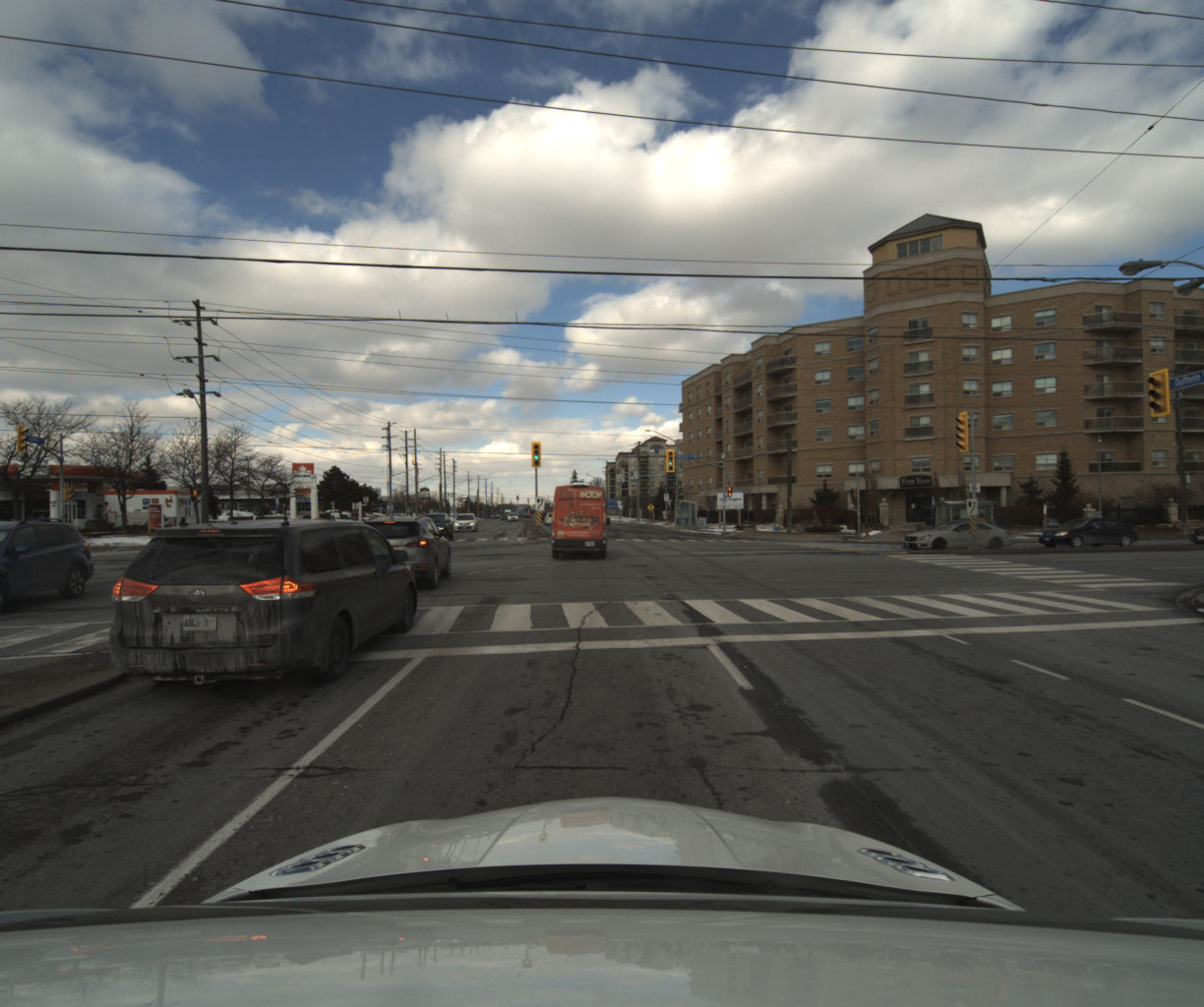}};
		\node[inner sep=0pt] (cam2) [right of=cam1, xshift=33mm]
			{\includegraphics[width=0.47\columnwidth]{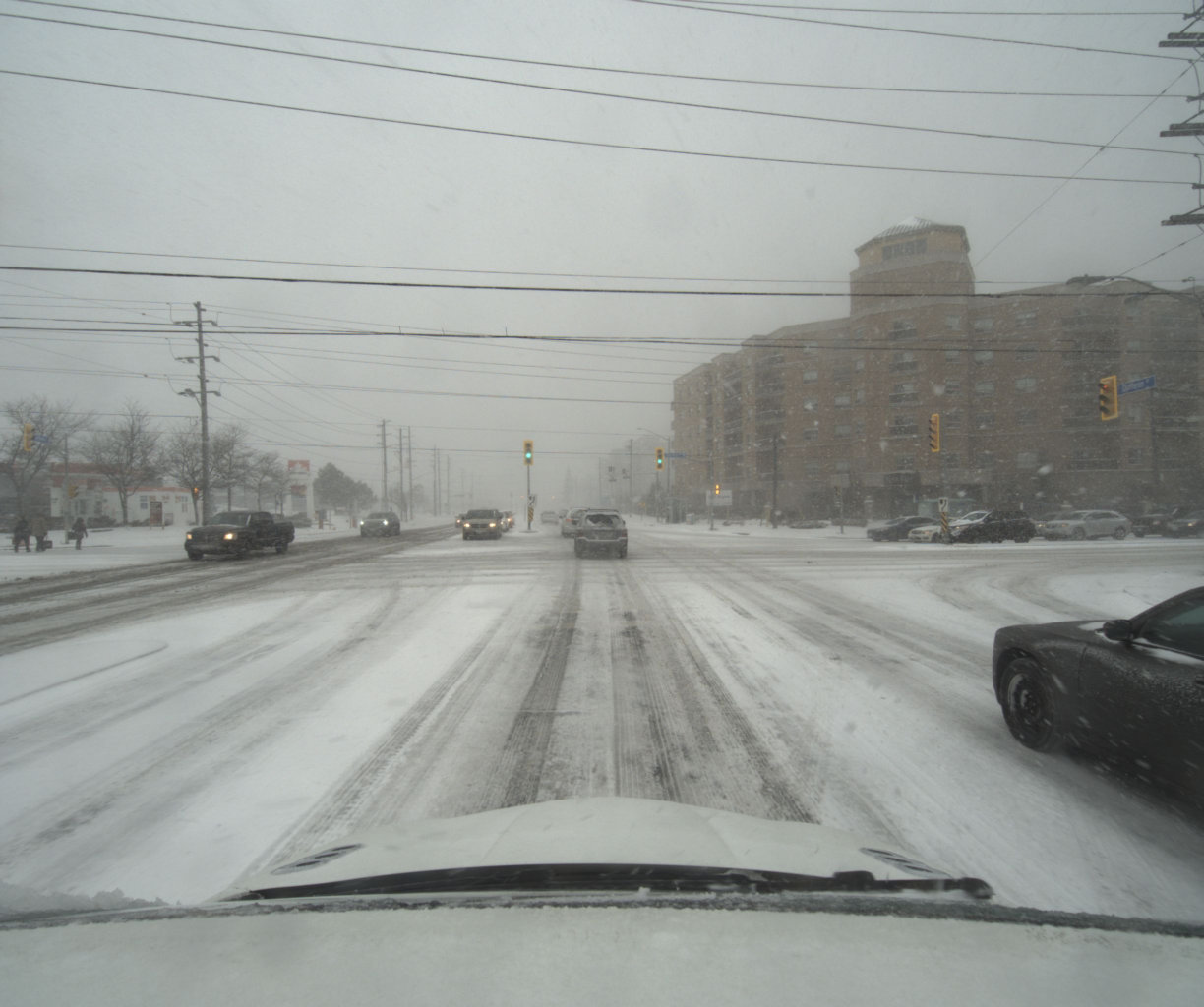}};
		\node[inner sep=0pt] (lidar1) [below of=cam1, yshift=-29mm]
			{\includegraphics[width=0.47\columnwidth]{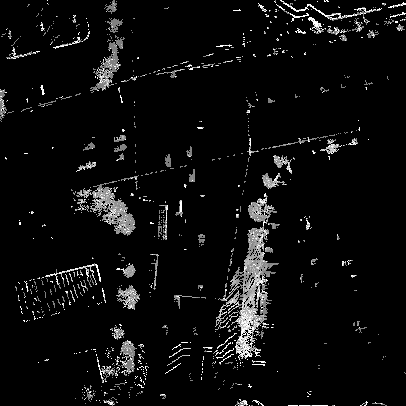}};
		\node[inner sep=0pt] (lidar2) [right of=lidar1, xshift=33mm]
			{\includegraphics[width=0.47\columnwidth]{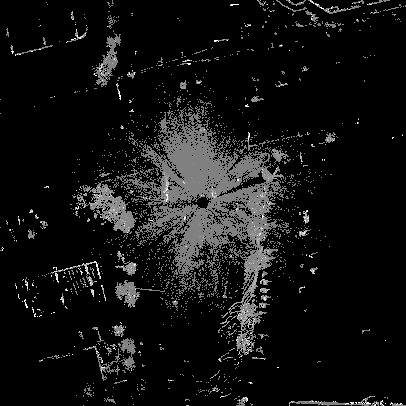}};
		\node[inner sep=0pt] (radar1) [below of=lidar1, yshift=-32.7mm]
			{\includegraphics[width=0.47\columnwidth]{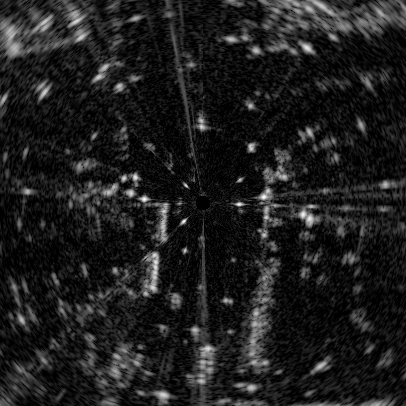}};
		\node[inner sep=0pt] (radar2) [right of=radar1, xshift=33mm]
			{\includegraphics[width=0.47\columnwidth]{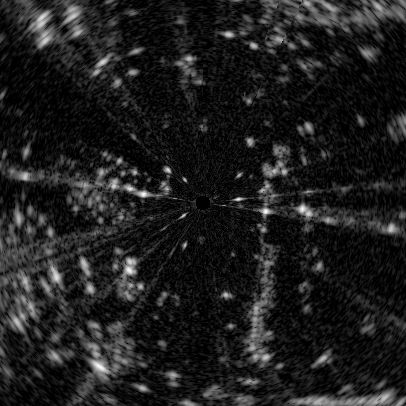}};
			
		\node (text1) [left of=cam1, xshift=-13mm, rotate=90] {\textbf{Camera}};
		\node (text2) [left of=lidar1, xshift=-13mm, rotate=90] {\textbf{Lidar}};
		\node (text3) [left of=radar1, xshift=-13mm, rotate=90] {\textbf{Radar}};
		\node (text4) [above of=cam1, yshift=9.5mm] {\textbf{Sun}};
		\node (text5) [above of=cam2, yshift=9.5mm] {\textbf{Snow}};
				
	\end{tikzpicture}
	\caption{This figure illustrates the differences in camera, lidar, and radar data between a sunny day and a snow storm. During snowfall, the lidar sensor becomes littered with noisy snowflake detections but the radar data appears unperturbed.}
	\label{fig:weather}
	\vspace{-6mm}
\end{figure}

Figure \ref{fig:weather} illustrates the large differences in weather conditions. Between the sunny and snowy days, there were significant visual appearance changes. Most of the lane lines are not visible during the snow storm. Even more stark is the contrast between the lidar data taken on the two days (after the ground plane is removed). During the snow storm, the lidar scan is littered with a large number of detections associated with snow flakes. Although we do not provide experimental results for this, it seems probable that the accuracy of a lidar odometry system would suffer under these conditions. However, \citet{charron_crv18} have shown that it is possible for these effects to be compensated. Unsurprisingly, it is difficult to discern the difference between the radar scans, other than the movement of large vehicles.
\vspace{-0.5mm}

\begin{figure}[t]
	\centering
	\includegraphics[width=0.90\columnwidth]{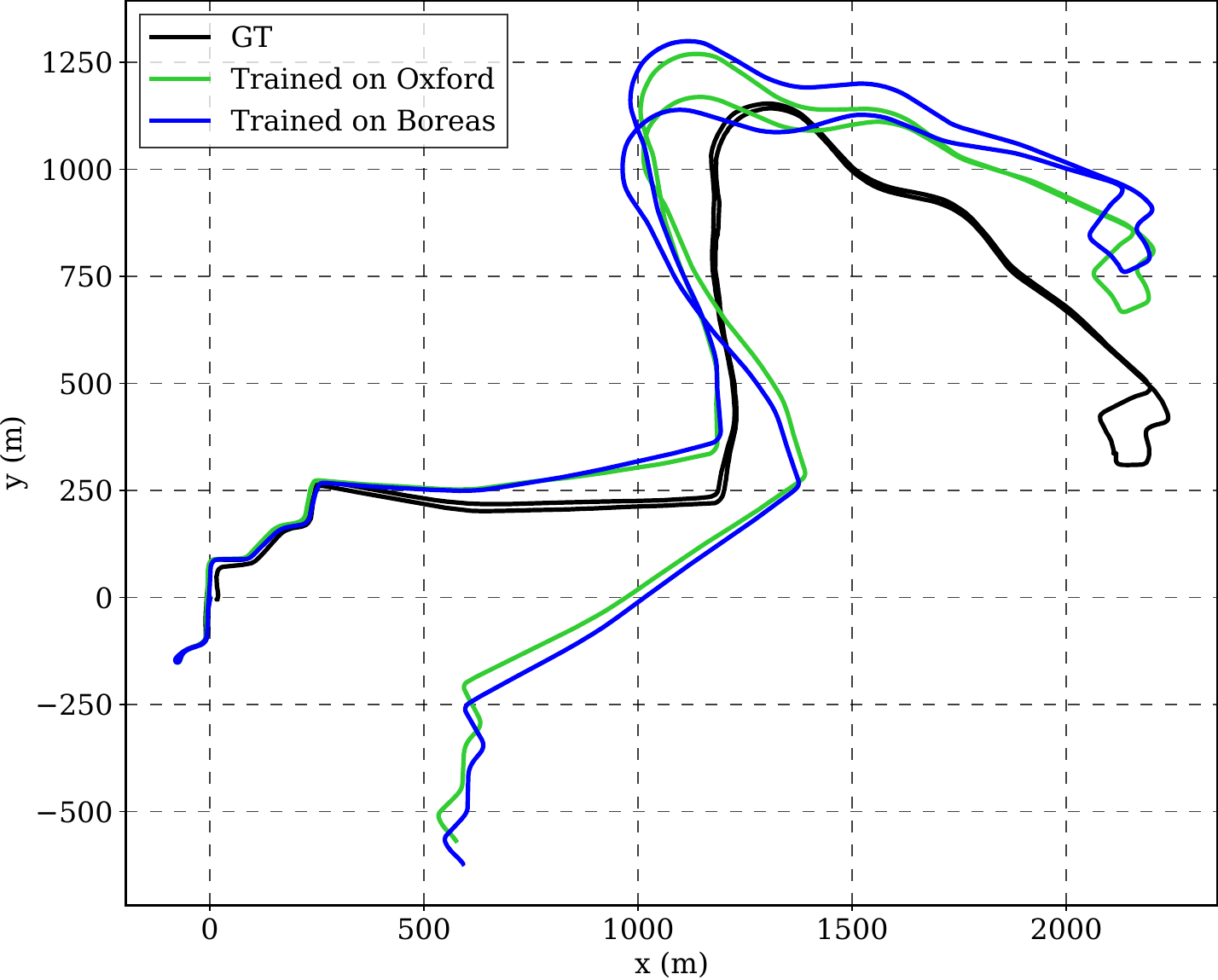}
	\caption{Odometry results on the snowy sequence 2021-01-26-11-22.\vspace{-2mm}}
	\label{fig:boreas_odom}
	\vspace{-2mm}
\end{figure}

The robustness of radar to inclement weather is also exemplified in the odometry drift results reported in Table~\ref{tab:boreas}. In fact, the drift rates on the snowy days were lower than on the sunny day. We postulate that the error increases on the sunny day due to the increase in the vehicle's speed. It is interesting to note that our rotation drift metrics here are lower than on the Oxford dataset. While it is difficult to point to a single factor as the reason for this difference in performance, we note that our trajectory involves less turns and we use a different source of groundtruth for testing. In Figure~\ref{fig:boreas_odom}, we provide a plot of our odometry results during a snowy sequence. Here, we have plotted the results of training on the Oxford dataset and testing on this sequence, as well as training on the \change{Boreas} dataset and testing on this sequence. There is a noticeable increase in the drift rate when transferring from the Oxford dataset to this \change{Boreas} dataset. We hypothesize that this difference is mainly due to the different range resolutions of the sensors.
\vspace{-1mm}


\renewcommand{\arraystretch}{1.1}
\begin{table}[t]
	\centering
	\caption{\change{Boreas} test results. \vspace{-2mm}}
	\begin{tabular}{|l|c|c|c|c|}
		\hline
		Sequence         & Weather & Trained On &{\begin{tabular}[c]{@{}c@{}}Trans.\\ Error (\%)\end{tabular}} & \multicolumn{1}{c|}{\begin{tabular}[c]{@{}c@{}}Rot. Error\\ (deg/1000m)\end{tabular}} \\ \hline
		01-26-10-59 & Snow & Boreas & 2.003                    & 5.599                                                                                           \\
		01-26-11-22 & Snow & Boreas & 1.980                    & 5.288                                                                                           \\
		02-09-12-55 & Sun & Boreas  & 2.073                    & 5.886                                                                                           \\ \hline
		01-26-10-59 & Snow & Oxford & 2.355                    & 8.137                                                                                           \\
		01-26-11-22 & Snow & Oxford & 2.112                    & 6.442                                                                                           \\
		02-09-12-55 & Sun & Oxford  & 2.546                    & 8.911                                                                                           \\ \hline
	\end{tabular}
	\label{tab:boreas}
	\vspace{-6mm}
\end{table}

\section{Conclusions and Future Work} \label{sec:conclusion}
\vspace{-1.5mm}
In this paper, we applied the \ac{ESGVI} parameter learning framework to learn radar features for odometry using only the on-board radar data. Our odometry performance on the Oxford Radar RobotCar Dataset approaches the current state of the art, which is a hand-crafted method \cite{adolfsson_arxiv21}. We provided additional experimental results on 100 km of driving taken in an urban setting. Within this dataset, we demonstrated the effectiveness of radar odometry during heavy snowfall.

\vspace{-0.5mm}

In recent years, robotics research has become increasingly reliant on high-quality datasets for training deep learning algorithms. The size and scope of these datasets is a limiting factor in the performance of many robotic systems. However, collecting and annotating these datasets is an expensive and labour-intensive process. For this reason, it is important that research in robotics trend towards solutions that are more data efficient. By foregoing the need for groundtruth pose information, our unsupervised architecture enables large quantities of training data to be collected with relative ease.

\vspace{-0.5mm}


Our framework is a hybrid of modern deep learning and classic probabilistic state estimation. Deep learning can be used to process rich sensor data while probabilistic estimation can be used to handle out-of-distribution samples through outlier rejection schemes. Furthermore, our modular framework enables many future extensions and avenues of research. For example, we plan on incorporating IMU factors into our estimator, which should be straightforward with our probabilistic estimator.


\newpage
\section*{Acknowledgments}

\small{We would like to thank Applanix Corporation and the Natural Sciences and Engineering Research Council of Canada (NSERC) for supporting this work. We personally thank Andrew Lambert at Applanix and Haowei Zhang at ASRL for helping with processing and collecting data. We thank General Motors for their donation of the Buick (vehicle).}


\bibliographystyle{plainnat}
\bibliography{bib/refs}

\newpage
\onecolumn
\appendix

\begin{table}[!ht]
	\centering
	\caption{Evaluation on 7 sequences from the Oxford Dataset. Performance is reported as translational drift (\%) / rotational drift (deg/1000m) using the common KITTI odometry metric. This table is restricted to algorithms that have been tested on similar sequences. Cen RO, Kung RO, and MC-RANSAC are hand-crafted algorithms. Masking by Moving, Under the Radar, and HERO are all learning-based algorithms. HERO is currently the only unsupervised learning-based radar odometry approach. *Tested on all 32 sequences. **Uses dense correlation between scans.\vspace{-3mm}}
	\label{tab:apptab1}
	\begin{tabular}{|l|c|c|c|c|c|c|c|c|c|}
		\hline
		\textbf{Method}   & \multicolumn{1}{c|}{\textbf{Evaluation}} & \multicolumn{7}{c|}{\textbf{Sequences}}                                         & \multicolumn{1}{c|}{\textbf{Mean}} \\ \hline
		&                                          & 10-14-02  & 11-12-26  & 11-14-02  & 14-12-05  & 15-13-06 & 16-11-53 & 17-11-46  &                                    \\ \hline
		Cen RO \cite{cen_icra18}            & \cite{barnes_icra20}                                         & N/A          & N/A           & N/A          & N/A          & N/A         & N/A         & N/A          & 3.72/9.5                           \\
		Kung RO* \cite{kung_icra21}            & \cite{kung_icra21}                                          & N/A          & N/A           & N/A          & N/A          & N/A         & N/A         & N/A          & 1.96/6.0                           \\ 
		Masking by Moving** \cite{barnes_corl19} & \cite{barnes_corl19}                                         & N/A          & N/A          & N/A          & N/A          & N/A         & N/A         & N/A          & 1.16/3.0                           \\ 
		Under the Radar \cite{barnes_icra20}   & \cite{barnes_icra20}                                          & N/A          & N/A          & N/A          & N/A          & N/A         & N/A         & N/A          & 2.05/6.7                           \\ 
		MC-RANSAC \cite{burnett_ral21}         & Ours                                         & 3.43/11.2 & 3.72/12.9 & 3.27/10.9 & 3.41/10.7 & 3.07/9.8 & 3.12/9.9 & 3.21/11.0 & 3.31/10.9          \\ 
		HERO (Ours)       & Ours                                         & 1.86/5.9  & 1.85/6.2  & 1.93/6.5  & 1.96/6.8  & 1.80/6.1 & 2.43/7.1 & 2.08/7.1  & 1.99/6.5                           \\ \hline
	\end{tabular}
\end{table}

\vspace{-4mm}

\begin{figure*}[!h]
	\centering
	\subfigure[11-12-26]{\includegraphics[width=0.30\columnwidth, clip]{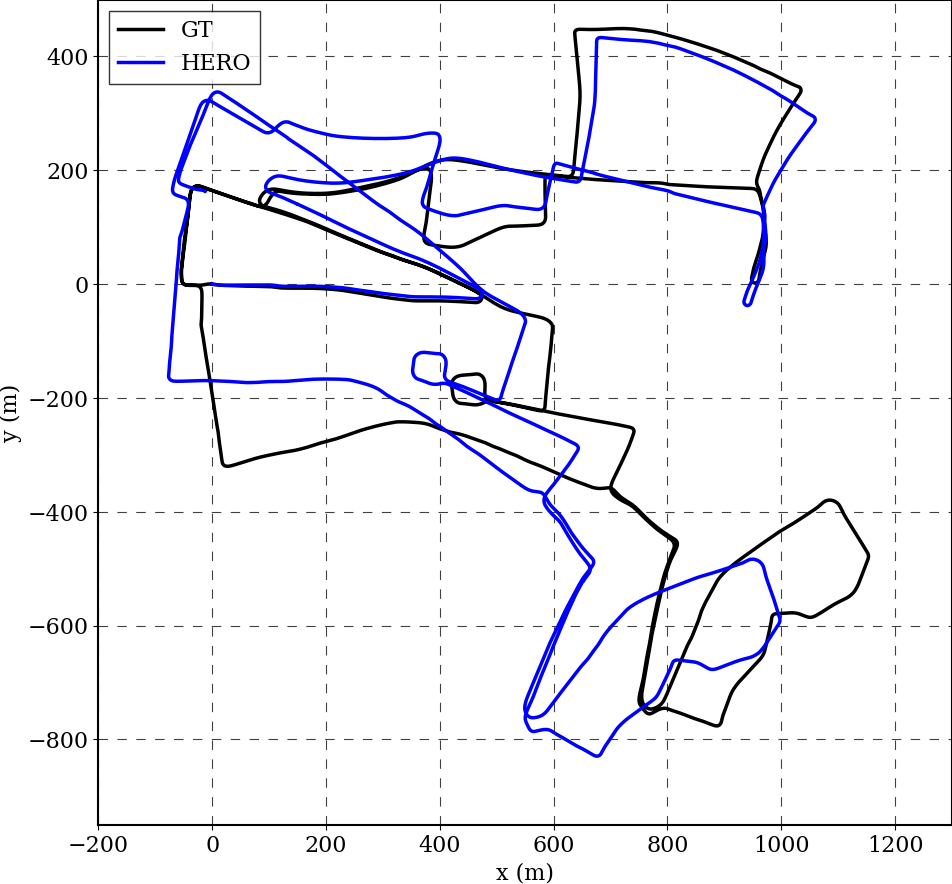}}
	\subfigure[11-14-02]{\includegraphics[width=0.30\columnwidth, clip]{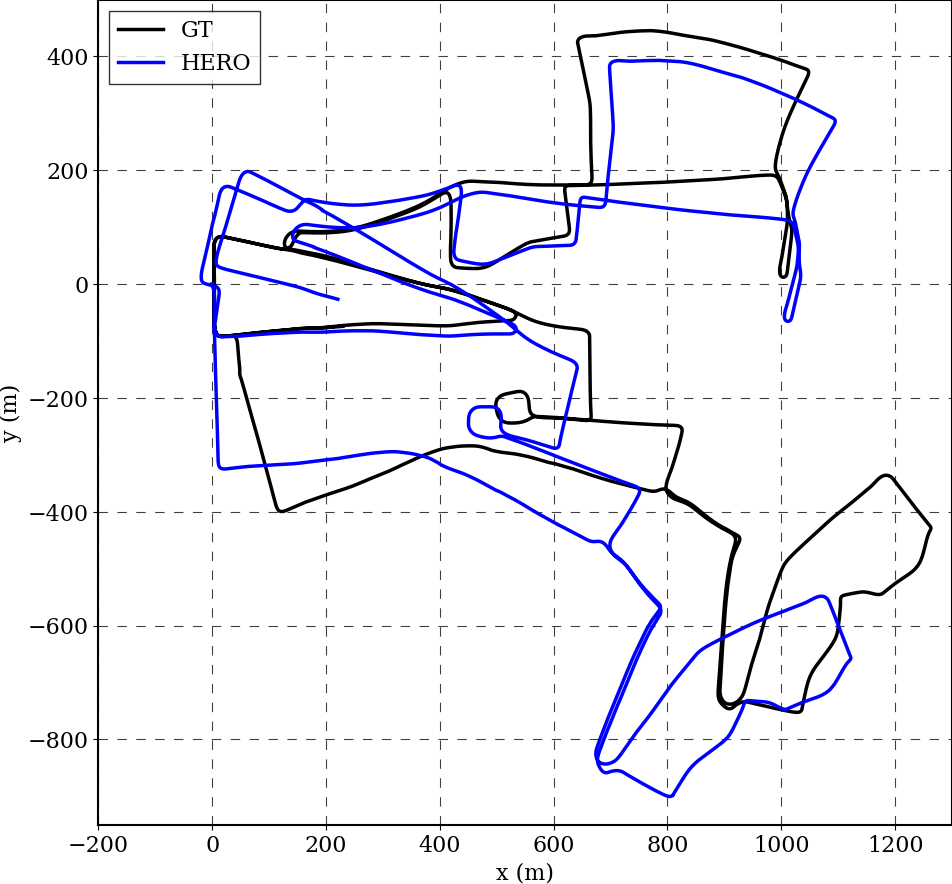}}
	\subfigure[14-12-05]{\includegraphics[width=0.30\columnwidth, clip]{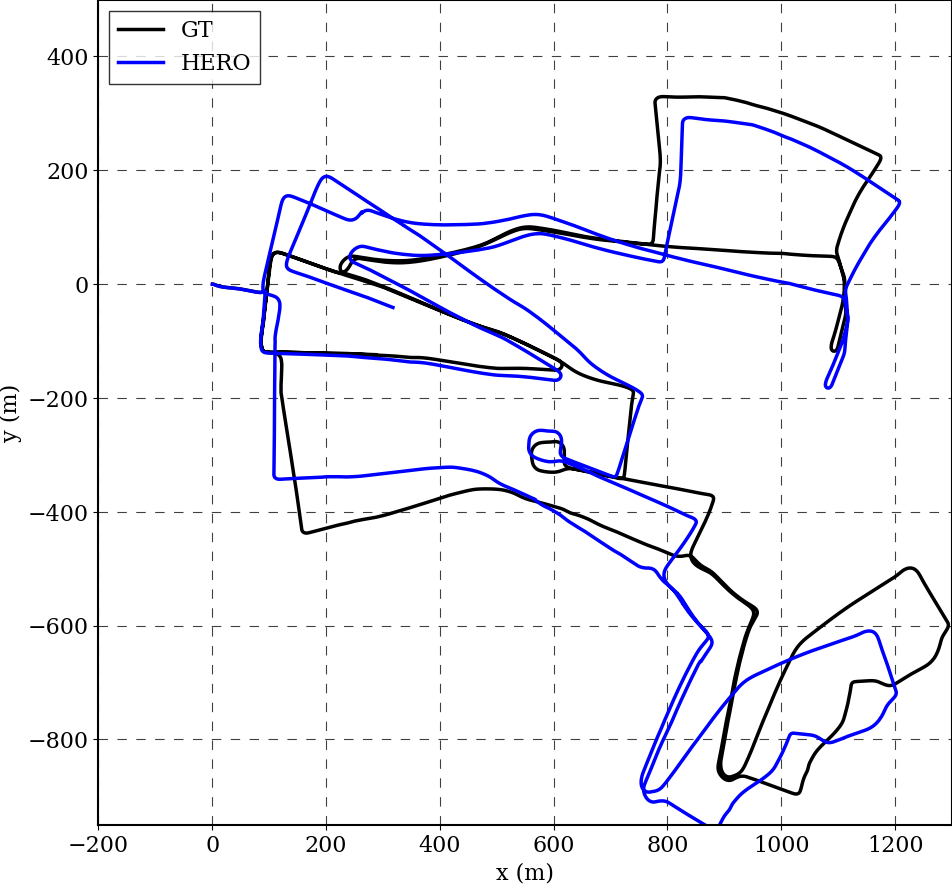}}
	\subfigure[15-13-06]{\includegraphics[width=0.30\columnwidth, clip]{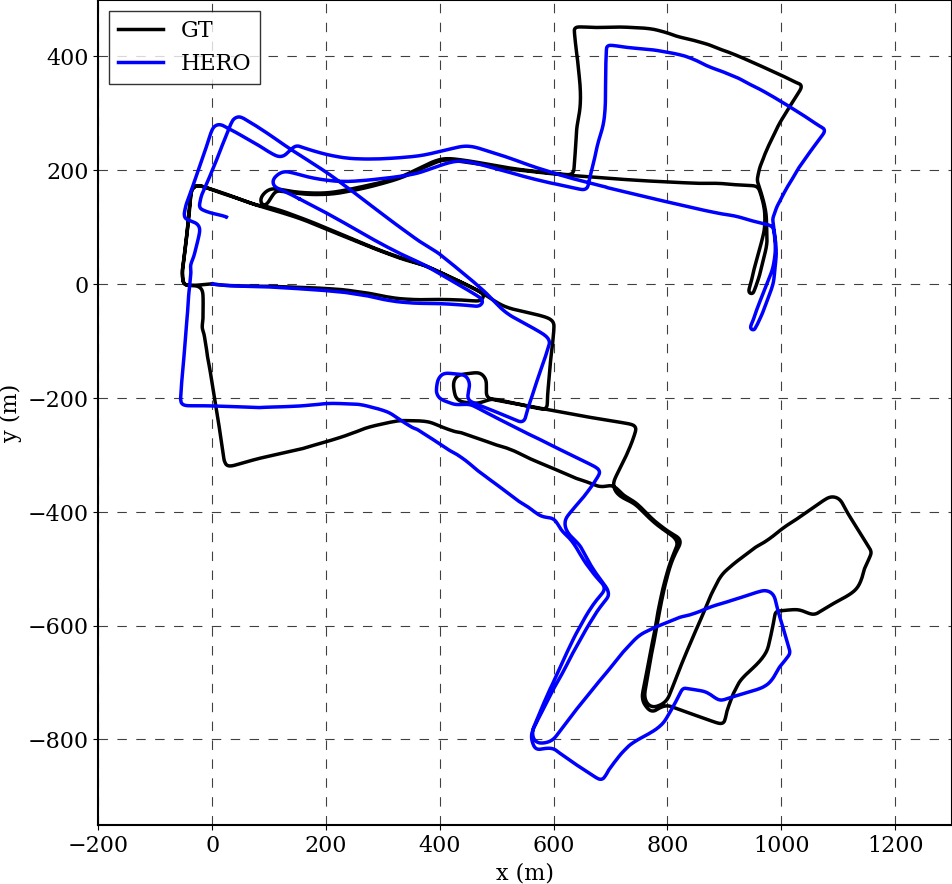}}
	\subfigure[16-11-53]{\includegraphics[width=0.30\columnwidth, clip]{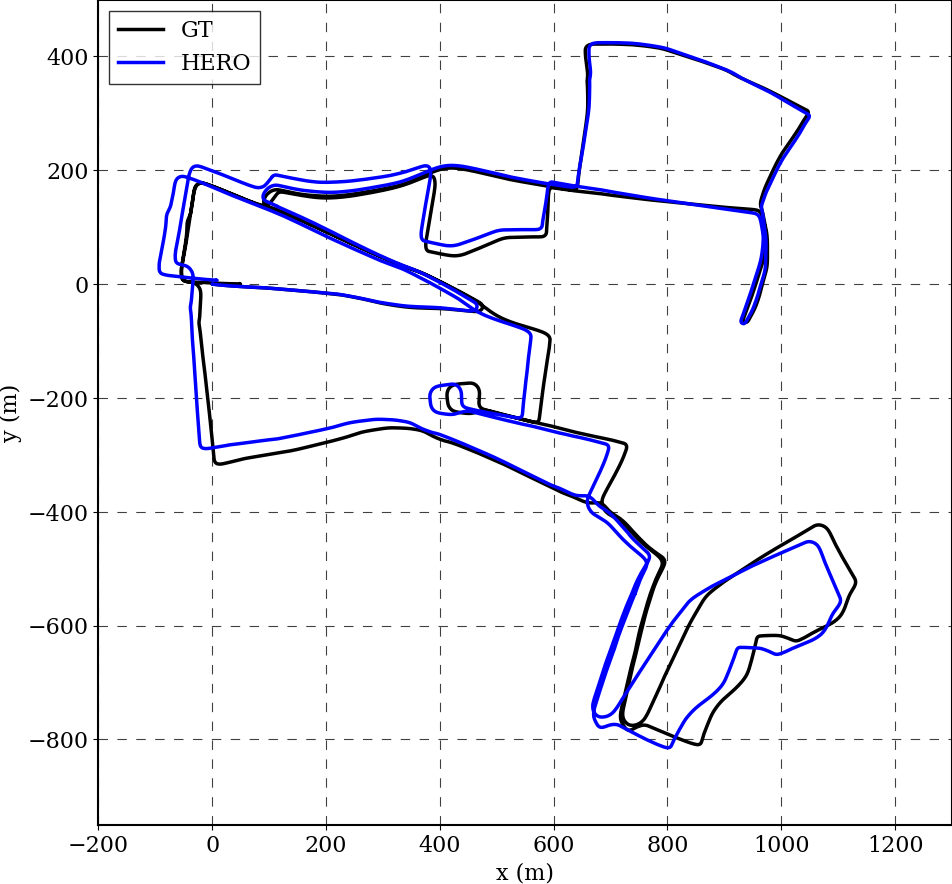}}
	\subfigure[17-11-46]{\includegraphics[width=0.30\columnwidth, clip]{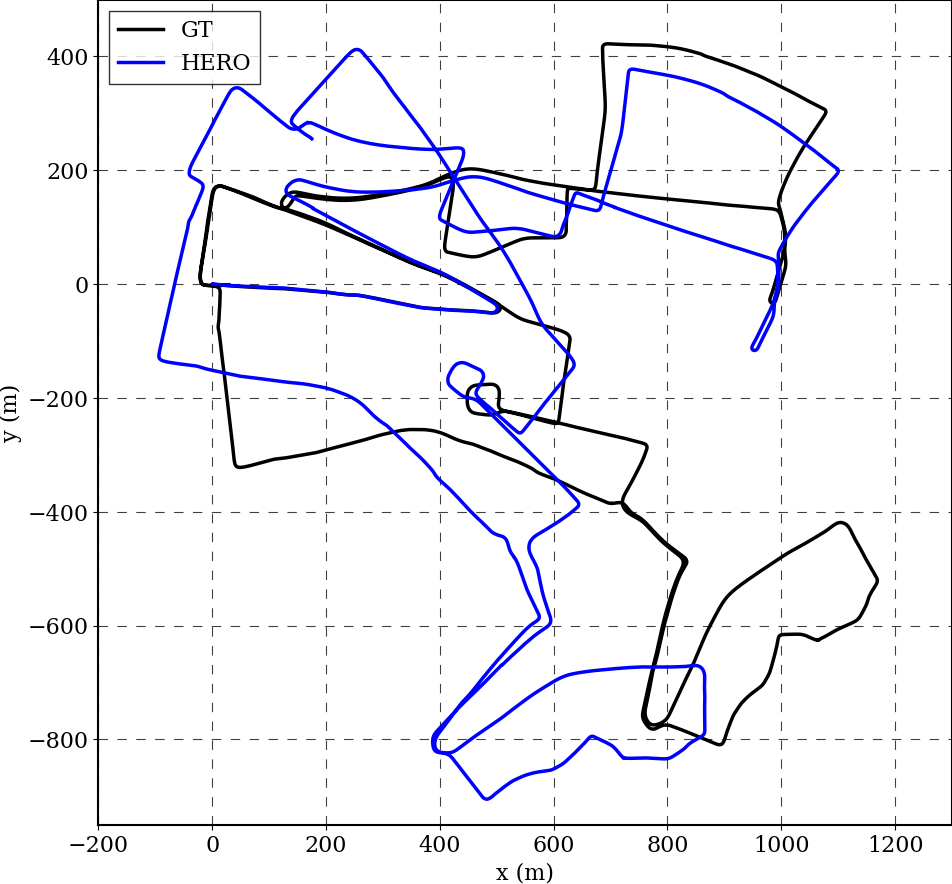}}
	\vspace{-3mm}
	\caption{These results illustrate the performance of HERO during each of the test sequences. See Figure~\ref{fig:odom} for HERO's performance on 10-14-02.\vspace{-3mm}} 
	\label{fig:odoms}
\end{figure*}

\begin{table}[!h]
	\centering
	\caption{An evaluation on 8 sequences from the Oxford Dataset. This experiment required a new train/validation/test split. 16-13-42 was used for validation. Performance is reported as translational drift (\%) / rotational drift (deg/100m) using the common KITTI odometry metric. *Uses loop closures.\vspace{-3mm}}
	\label{tab:apptab2}
	\begin{tabular}{|l|c|c|c|c|c|c|c|c|c|c|}
		\hline
		\textbf{Method}   & \multicolumn{1}{c|}{\textbf{Evaluation}} & \multicolumn{8}{c|}{\textbf{Sequences}}                                         & \multicolumn{1}{c|}{\textbf{Mean}} \\ \hline
		&                                          & 10-11-46  & 10-12-32  & 16-11-53  & 16-13-09  & 17-13-26 & 18-14-14 & 18-14-46 & 18-15-20  &                                    \\ \hline
		Hong RO \cite{hong_arxiv21}            & \cite{hong_arxiv21}                                        & 2.16/0.6          & 2.32/0.7           & 2.49/0.7          & 2.62/0.7          & 2.27/0.6         & 2.29/0.7         & 2.12/0.6          & 2.25/0.7 & 2.32/0.7                           \\ 
		Hong SLAM* \cite{hong_arxiv21} & \cite{hong_arxiv21}                                        & 1.96/0.7          & 1.98/0.6          & 1.81/0.6          & 1.48/0.5          & 1.71/0.5         & 2.22/0.7         & 1.68/0.5          & 1.77/0.6 & 1.83/0.6                           \\ 
		CFEAR \cite{adolfsson_arxiv21}   & \cite{adolfsson_arxiv21}                                          & 1.65/0.48          & 1.64/0.48          & 1.99/0.53          & 1.86/0.52          & 1.66/0.48         & 1.71/0.49         & 1.79/0.5          & 1.75/0.51 & 1.76/0.50                           \\ 
		HERO (Ours)       & Ours                                         & 2.14/0.71  & 1.77/0.62  & 2.01/0.61  & 1.75/0.59  & 2.04/0.73 & 1.83/0.61 & 1.97/0.65  & 2.20/0.77 & 1.96/0.66                           \\ \hline
	\end{tabular}
\end{table}

\end{document}